\documentclass[12pt]{article}

\usepackage{url}
\usepackage{scicite}
\usepackage{comment}

\usepackage{times}
\usepackage{xcolor}
\usepackage{graphicx}

\newcommand*\samethanks[1][\value{footnote}]{\footnotemark[#1]}

\newenvironment{sciabstract}{%
\begin{quote} \bf}
{\end{quote}}

\title{Socially Cognizant Robotics\\ for a Technology Enhanced Society} 
\date{}
\author{Kristin J. Dana\thanks{Electrical and Computer Engineering Dept.},
Clinton Andrews\thanks{Bloustein School of Planning and Policy}, 
Kostas Bekris\thanks{Computer Science Dept.},
Jacob Feldman\thanks{Psychology Dept.}, 
Matthew Stone\samethanks[3],\\
Pernille Hemmer\samethanks[4],
Aaron Mazzeo\thanks{Mechanical and Aerospace Engineering},
Hal Salzman\samethanks[2], 
Jingang Yi\samethanks[5]\\Rutgers University}

\begin{document} 


\baselineskip24pt


\maketitle

% The preamble here sets up a lot of new/revised commands and
% environments.  It's annoying, but please do *not* try to strip these
% out into a separate .sty file (which could lead to the loss of some
% information when we convert the file to other formats).  Instead, keep
% them in the preamble of your main LaTeX source file.

\begin{sciabstract}
Emerging applications of robotics, and  concerns about their impact, require  the research community to put human-centric objectives front-and-center.  To meet this challenge, we advocate an interdisciplinary approach, {\it socially cognizant robotics}, which synthesizes technical and social science methods. We argue that this approach follows from the need to empower stakeholder participation (from synchronous human feedback to asynchronous societal assessment) in shaping AI-driven robot behavior at all levels, and leads to a range of novel research perspectives and problems both for improving robots' interactions with individuals and 
%for improving robots' 
impacts on society. Drawing on these arguments, we develop best practices for {\it socially cognizant robot design} 
%that balance traditional technology-based metrics (e.g.\ efficiency, precision and accuracy) with critically important objectives, albeit challenging ones to measure, describing the effects of robots on individuals and on society.
that balance traditional technology-based metrics (e.g.\ efficiency, precision and accuracy) with critically important, albeit challenging
to measure, human and society-based metrics.

%With the growing potential of robotics, as well as concerns of its impact, we consider {\it socially cognizant robotics} for the interdisciplinary  study of robotics and humanity. In this article we address the questions-- What is socially cognizant robotics? Why is it timely and beneficial?  We propose that {\it human engagement affecting robot design} is a key component of socially cognizant robotics and identify research challenges that coalesce into two major themes: 1) the human-robot dyad and 2) the robot-society dyad. Motivated by a set of overarching goals for this multi-faceted discipline, we develop best practices for {\it socially cognizant robot design} to balance traditional technology-based metrics (e.g.\ efficiency, precision and accuracy) with critically important,  albeit challenging to measure, human and  society-based metrics. 
\end{sciabstract}

\section{Introduction}
\label{sec:intro}
Applications of robotics (such as  telepresence, transportation, elder-care, remote health care, cleaning, warehouse logistics, and delivery) are bringing significant changes in individuals’ lives and are having profound social impact. Despite the envisioned potential of robotics, the goal of ubiquitous robot assistants augmenting quality of life (and quality of work life) has not yet been realized.  Key challenges lie in the complexities of {\bf four overarching human-centric objectives} that such systems must aim for: 1) improving quality of life of people, especially marginalized communities; 2) anticipating and mitigating unintended negative consequences of technological development; 3) enabling robots to adapt to the desires and needs of human counterparts;  4) respecting the need for human autonomy and agency. 

\begin{figure}
    \centering
    \includegraphics[width=5.8in]{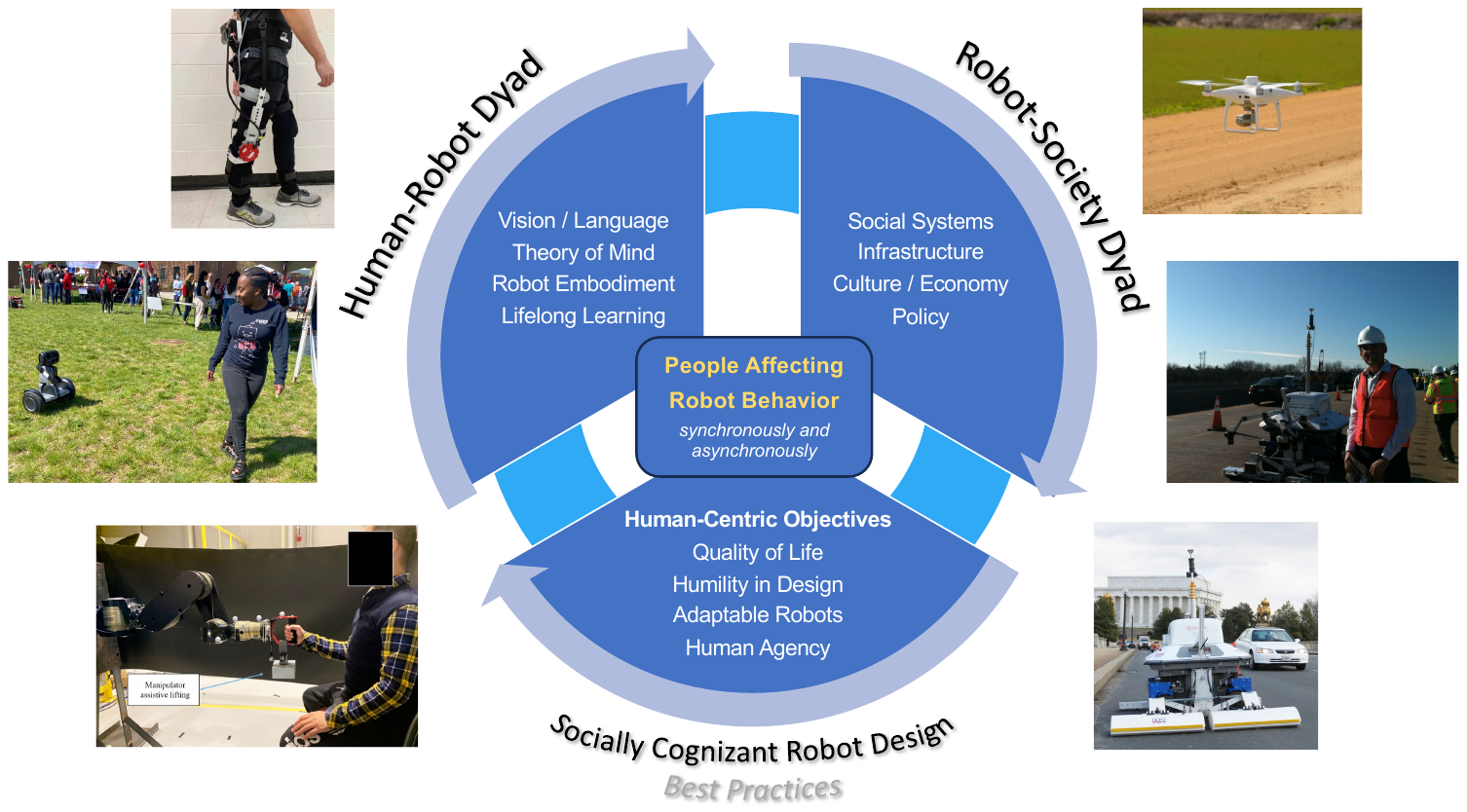}
    \caption{
      Considerations of the human-robot dyad and the robot-society dyad within this nascent interdisciplinary field of socially cognizant robotics lead to a design framework based on human-centric objectives. 
    Empowering AI stakeholder participation (ranging from  individual feedback  to societal assessment) ensures that people can {\it affect} and tune robot behavior.  This participation can be synchronous (real-time) or asynchronous as described in Figure~\ref{fig:DyadFigure}. }
    \label{fig:scrobotics_overview}
\end{figure}

Pursuing these objectives requires an integrated cohort of technologists, behavioral scientists and social scientists with a shared vision to pursue a deep, multidisciplinary understanding of how robots interact with individuals and society.  We introduce a new term, {\bf socially cognizant robotics}, to describe this multi-faceted interdisciplinary branch of technology. The emerging practitioner, the socially cognizant roboticist, represents the convergence of {\it socially aware technologists}, who can develop intelligent devices that adapt to human and social behavior; and {\it technology-aware social scientists and policymakers}, who can translate studies of robotics’ social effects into actionable and technically-viable principles and policies.

A primary element of {\it socially cognizant robotics} is a deliberate ``invitation to the table'' for social scientists, who bring analytical perspectives and methods that are not typically present in robotics.  These perspectives cover two levels of human-technology interaction that we view as  essential: the {\bf human-robot dyad} 
{\it (Section~\ref{sec:humanrobotdyad})}
and the {\bf robot-society dyad}
{\it (Section~\ref{sec:robotsocietydyad})}. 
Figure~\ref{fig:scrobotics_overview} illustrates how these levels might operate in the context of the workplace and everyday life. 
%The overall aim is to 
These considerations lead to formulating {\bf best practices in socially cognizant robot design} {\it (Section~\ref{sec:SCdesign})} that emphasizes the role of feedback (from individual and societal stakeholders) so that humans can affect and tune robot behavior.

\section{Human-Robot Dyad}
\label{sec:humanrobotdyad}

How can robots effectively interact with human partners? Technologists are making great strides in supporting intelligent interaction in fields such as {\bf robot embodiment, control and planning} \cite{brohan2022rt,vemprala2023chatgpt, shah2023lm}, {\bf visual learning} \cite{radford2021learning,dosovitskiy2020image,liu2021swin,dehghani2023scaling}, and {\bf language processing} \cite{kenton2019bert,brown2020language,vaswani2017attention}. But excitement due to the technological progress is tempered by an appreciation of the limitations and dangers of current methods \cite{raji2ai,bender2021dangers,thorp2023chatgpt,gebru2021datasheets}, especially for data-driven approaches that control action in the physical world in the presence of people without accompanying safety guarantees.  

Findings and methods from {\bf psychology and cognitive science} suggest how some of these limitations may be addressed. For example, the ability to understand and anticipate human needs and desires is often referred to in psychology as “theory of mind” (ToM) \cite{Leslie_TOM}; ToM provides a psychologically-informed framework for designing robotic behavior that mirrors human--human interaction. Such ideas might be realized through developing and employing human-in-the-loop approaches to construct human informed training sets, as well as to guide robot activity in real time.  %Another direction is to use 
Additionally, cognitive models may identify and meet the needs of the individual under real-world circumstances that are out-of-distribution during lab testing.

%Emphasizing human engagement during robot training is an acknowledgment of the limitations and dangers of developing robotic technology separated from the needs and desires of the people who will be interacting with the resulting technology in the real world.

  %Datasets derived from human interaction are also important for robot training  (e.g.\ image annotations or videos of human action), but this offline interaction is distinct from real-time online engagement for shaping robot behaviors. 

%
%%%%%% Theory of Mind + Advanced Embodiment
%%%%%%%%%%%%%%%%%%%%

%\subsection{Vision/Language/Action Models Support Human Interaction}
\subsection{From Vision and Language to Robot Action}

Progress in computer vision (CV)
and natural language processing (NLP) has made it possible for robots to perceive and represent their environment, including the humans within it.
Both CV and NLP are faced with the challenge of extracting meaningful inferences and communication from raw environmental stimuli---the same challenge solved by the human brain.  Computationally, raw  signals must be converted to suitable intermediate representations in order to extract knowledge. For example, simple image distance metrics are not usually useful, but can be the basis of more structured and meaningful measurements of similarity \cite{destler_singh_feldman}.  This challenge of  representation learning is fundamental in both CV and NLP research. 
In CV, CNNs have become a popular architecture due to their robust performance \cite{krizhevsky2012imagenet,lecun1995convolutional}.
More recently the transformer architecture in NLP has greatly improved language translation and generation \cite{vaswani2017attention,devlin2018bert}.
The transformer architecture has been repurposed for visual recognition and now largely outperforms CNNs. Recent demonstrations of transformer-based models generating text (e.g., ChatGPT 
\cite{brown2020language, radford2018improving,radford2021learning})
and images (e.g., Dall-E) \cite{ramesh2021zero}) 
have led to a surge of interest in these frameworks. 
Current CV and NLP frameworks have uncovered computational representations, useful for understanding and generating both images and language, in a manner that is reminiscent of human ability (although with significant limitations), and the implications for robotics is significant.
These frameworks from CV and NLP are inspiring recent similar models for robot action or control  
\cite{zhou2019does, brohan2022rt, vemprala2023chatgpt, reed2022generalist} 
including new approaches of reinforcement learning
\cite{Chen2021decisiontransformer,janner2021offline}.
The promise of these methods are tempered by inherent dangers: \cite{floridi2020gpt, bommasani2021opportunities}:
mistakes in actions models can do more immediate harm, making a socially cognizant approach to robotics particularly timely and urgent.  

\begin{figure}
    \centering
    \includegraphics[width=6.1in]{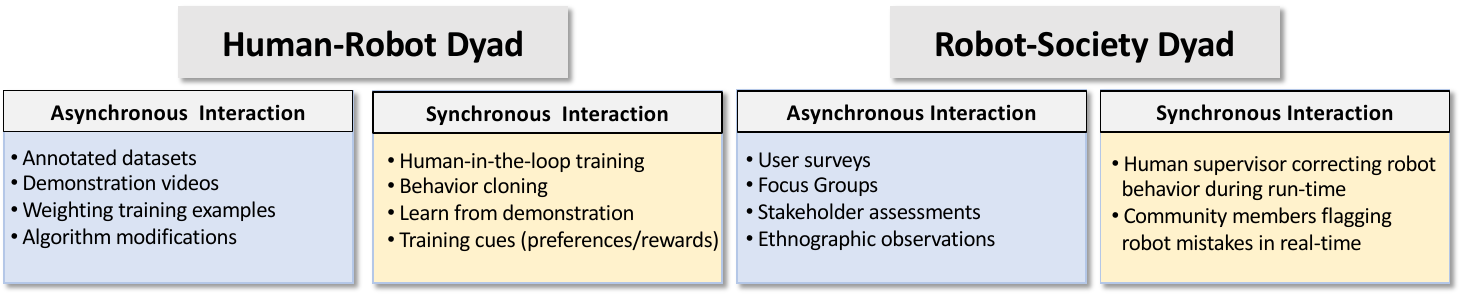}
    \caption{Learning from feedback: A socially cognizant
robot should interact with and adapt to humans
both synchronously and asynchronously within the
human-robot dyad or a society-robot dyad. }
    \label{fig:DyadFigure}
\end{figure}

\subsection{Human-in-the-Loop Robot Training}
%% Note: modify this to be 1-2 paragraphs but include RLHFA 

In socially cognizant robotics, humans are not merely part of the robot's environment (e.g.\ to consider for collision avoidance). Rather, the human-in-the-loop meaningfully participates in shaping robot behavior and robot training. 
Socially cognizant robotics empowers AI stakeholder participation at multiple levels  (such as individuals interacting with robots, expert dataset annotators, user feedback, and  group assessment surveys) both synchronously and asynchronously, as shown in Figure~\ref{fig:DyadFigure}, with the goal of improving the outcomes of robot interactions.
%(see figure). 

A common and well-developed method for humans training robots is through supervised machine learning where datasets are annotated by humans  (including our research \cite{zhang2018context,Xue2017,xue2018deep,la2017development,akiva2022vision}).
This asynchronous human interaction takes the form of large labeled datasets where humans have provided the knowledge label (e.g.\ labeling the contents of an image or the location of an object in the image) for training computer or robot vision perception tasks such as image recognition, semantic segmentation, object detection, instance segmentation and object counting. 
For robot control, this asynchronous human input includes video observations of task demonstration that can be used to learn human behavior \cite{pearce2023imitating, shafiullah2022behavior}.
Problems associated with using datasets \cite{bender2021dangers, gebru2021datasheets, denton2021whose, koch2021reduced, raji2ai} 
include the cost and human effort of obtaining new datasets, ethical considerations for human annotators, problematic domain shifts, and a hyperfocus of research on obtaining results on existing datasets.  Anticipating and mitigating these negative consequences are an important part of the early design stage as discussed in Section~\ref{sec:SCdesign}.

A trend that is emerging from the need for human feedback is synchronous or online human-in-the-loop robot training. This is a paradigm of humans improving robots by interacting and intervening, i.e. {\it humans teaching robots}. 
Recent work in human-in-the-loop reinforcement learning provides examples of this paradigm \cite{wang2022skill, jang2022bc, lee2021pebble, chen2021ask,zhang2019leveraging, khalid2020combining}.  Human-in-the-Loop RL typically works by modifying the reward function, e.g. by indicating preferences for a particular robot behavior. A related concept is  imitation learning and learning by demonstration, which  \cite{lazaro2019beyond,shao2021concept2robot}. Recent hybrid methods use both asynchronous demonstration and intervention \cite{jang2022bc,schmeckpeper2021reinforcement}. Because such methods can be adjusted by humans during the operation,  humans can affect the design iterations and this paradigm is well aligned with the overarching goal of respecting human agency.  Additionally, human-in-the-loop methods may offer efficient performance improvements with less reliance on datasets.

\subsection{ Robot Theory of Mind}

One striking difference between conventional robots and human agents is that only the latter possess \textit{theory of mind}, the ability to understand other agents' mental states \cite{Leslie_TOM}. People attribute mental attributes---beliefs, intentions, and knowledge---to other agents almost reflexively \cite{Heider_Simmel,Blythe_Todd_Miller,Tremoulet_Feldman_2006}, utilizing a neural pathway apparently specialized for social relations \cite{Pitcher_Ungerleider_2021}. Indeed mental state attributions are essential for normal social interaction, especially when language is involved \cite{deVilliers_2007}: it is scarcely possible to interpret an instruction, a question, or even a single word without comprehending the speaker's likely state of knowledge. Joint action by multiple humans is similarly facilitated by people's ability to understand each other's mental processes \cite{Sebanz_etal_2006}. Indeed many of our inferences, actions, and reactions are predicated on an intuitive sense of what other people are thinking, what they want, and how they will likely perceive our own actions \cite{Gergely,Johnson_TICS}.
 
 It has long been speculated that making robots engage in human-like social interaction would require endowing them with ToM (e.g. \cite{Dautenhahn_2007}). 
 But progress on this problem has been slowed by our limited understanding of how human “theory of mind” actually works. Current models from our research and others \cite{Pantelis_etal_2014,Baker_etal_2017,JaraEttinger_Schulz_Tenenbaum_2020} are based on Bayesian estimation of other-agent mental states, allowing the model to guess the goals and desires of another agent in order to fully understand their actions and anticipate their needs. However, these models only work in very limited contexts, based on a small set of predetermined possible goals. In order to interact effectively with a person, a robot must be able to decide among hundreds or thousands of potential human desires, corresponding to the myriad goals a human user might have in interacting with an automated device \cite{Thellman_etal_2022}. An important example comes from navigation. People navigate around other people in part by using ToM to understand and anticipate the likely movements of others \cite{Dalton_Montello_2019, destefani2020spatial,johnson2023learning}. For a robot to navigate among humans in a similar manner would require a similar degree of social awareness \cite{Moller_etal_2021}. 
 Computational models of human ToM have been proposed  \cite{Pantelis_etal_2014,Baker_etal_2017}
and introduced into robotics \cite{DBLP:conf/iclr/PuigSLWLTF021}. 
Recent computational models for ToM  use deep learning \cite{rabinowitz2018machine} and inverse reinforcement learning \cite{jara2019theory}.
we expect computational ToM models to benefit from recent machine learning paradigms and architectures, such as transformers and human-in-the-loop RL. To achieve truly socially cognizant robotics,  such models need to include mechanisms for humans to synchronously and/or asynchronously affect and tune computational ToMs in order to affect robot behavior.

\begin{figure*}
    \centering
    \includegraphics[width=4.7in]{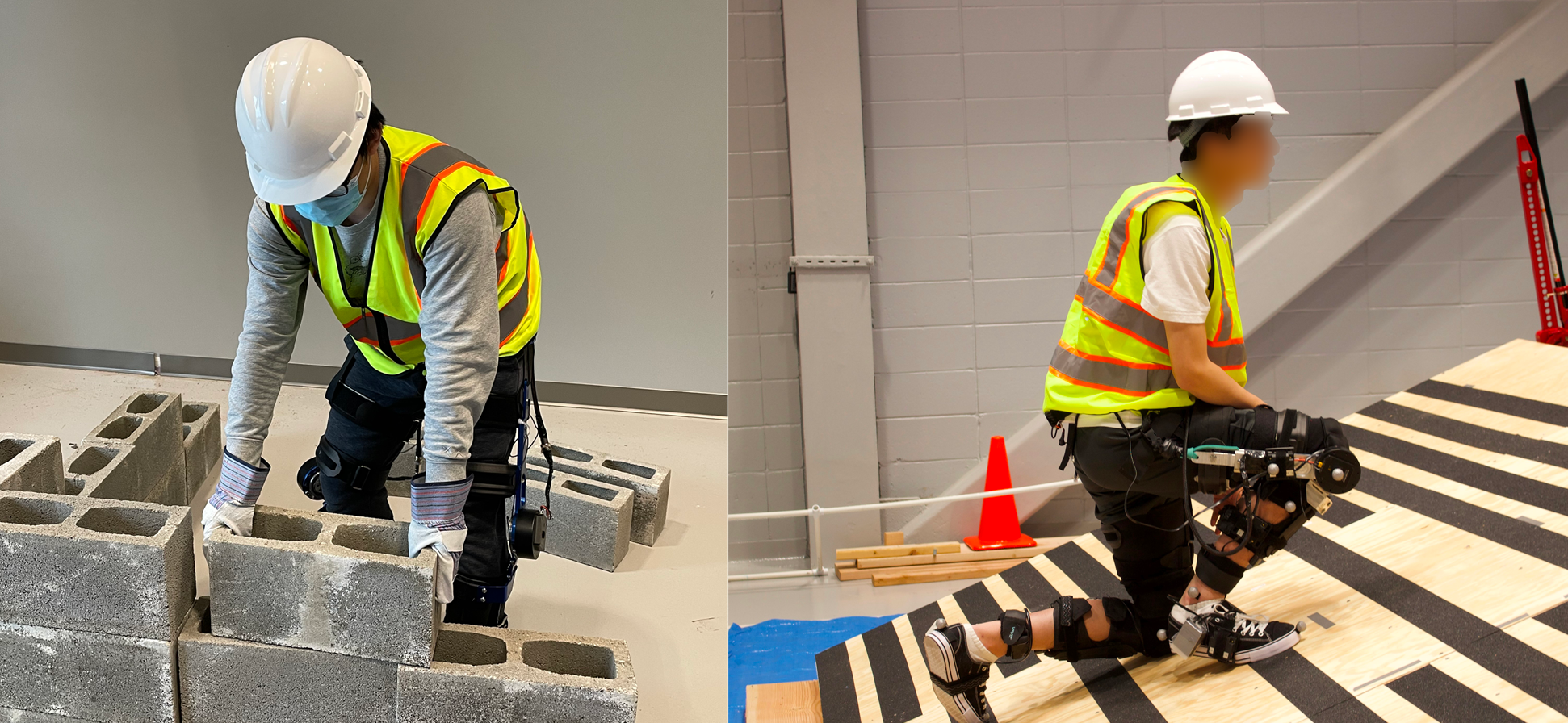}
    \caption{Modern robot embodiment includes exoskeletons.  Flexible, light-weight, high-torque bilateral knee exoskeletons are shown here in the task of assisting construction workers to reduce fatigue and injuries. Left: Masonry work. Right: Roof worker with kneeling gait \cite{chen2021wearable}.}
    \label{fig:exo}
\end{figure*}

\subsection{Robot Embodiment for Human Interaction}

Robots have thrived in highly structured, accurately known, and safely enclosed industrial
settings. Typically, these platforms consist of rigid elements connected by a few degrees of freedom (DoF) that can be controlled directly. This allows strict enforcement of motion for high speed, accuracy, and consistency of repeated tasks. In contrast, modern robotics aspires to more general and ambitious operational goals involving highly unstructured environments, arbitrary terrains, unknown objects, or more importantly in the context of Socially Cognizant Robotics, the company of humans. These altered priorities lead to the key challenge of robots with versatile physical traits, such as deformation and heightened articulation \cite{rus2015design}, which can be achieved by using soft components or a high number of compliant DoF \cite{polygerinos2017soft}. These features
allow for adaptive contact geometry and storage or dissipation of energy for purposes, such as efficient mobility on rough terrain and manipulation of arbitrary objects. Examples include soft \cite{deimel2016novel, gu2021soft} and adaptive hands \cite{sintov2019learning}, grippers \cite{ilievski2011soft,sinatra2019ultragentle}, manipulators \cite{george2018control}, locomotors [\cite{shepherd2011multigait}], skin-like sensors \cite{shih2020electronic,zou2018paper}, and hybrid soft-rigid tensegrity robots \cite{shah2022tensegrity}. Pioneering early work from our research and others in soft robotics \cite{ilievski2011soft,shepherd2011multigait,cutkosky2009design} seeded exponential growth in the field in recent years \cite{hawkes2021hard}. Robots with versatile physical traits also include soft robotic exoskeletons to provide strength augmentation for purposes of mobility assistance, rehabilitation and worker protection during repetitive lifting \cite{yap2015soft,yap2016high,xiloyannis2021soft, yu2020exo,chen2021wearable} as illustrated in Figure~\ref{fig:exo}. 
Socially cognizant robotics considers how technology, such as soft robotic material, can change the perceived empathy, competence, and safety of robots. This requires  integrated treatment of human intent, perception, and behavior in interaction with embodied and intelligent robots.

\subsection{Life-Long and Continual Learning with Memory}

 Robots of the future may  be trained by a combination of cognitive models, pre-trained large networks, human interaction, synthetic environments, and self-supervision from videos.  Robot devices, like individuals, are a unique combination of their training experiences, combining data, models and interactions. 
 The goal of life-long learning for intelligent agents \cite{de2021continual,parisi2019continual,tessler2017deep} necessitates addressing memory, deciding how and what to remember, so that training based on past experience can to be efficient.  
%Learning in AI neural networks
A significant shortcoming of artificial neural networks is that they experience catastrophic forgetting. While they can be efficiently trained  on a single task across a wide variety of domains, when trained on a new task they often fail to retain previously learned tasks. This is a fundamental challenge in life-long learning.
In contrast, humans, despite the very complex cognitive processes involved, efficiently encode and retrieve memories across the lifespan, without forgetting previously learned experiences. Human episodic memory is a reconstructive process, and one can think of episodic memory as a problem of extracting and storing information from the noisy signals presented to our senses, with the goal of efficiently storing and retrieving relevant information. Because the encoded memory traces are inherently noisy and incomplete reconstruction at recall requires input from at least two memory structures: one based on specific prior episodes, and one that is an abstraction of relevant prior knowledge and expectations (aka semantic memory). In this way, reconstructive memory can exploit environmental regularities to ‘clean up’ the noise in our memory system and improve average recall performance. (Importantly, this also prevents catastrophic forgetting.)  If an episodic trace is too noisy or inaccessible, without semantic memory the best the system can do is guess randomly and at worst have catastrophic forgetting. However, the optimal behavior is a trade-off between the strength of the evidence in episodic memory and the likelihood of the event based on prior knowledge and expectations. 
%Memory, and therefore learning, is context dependent. Furthermore, this influence of prior knowledge is hierarchical, such that prior knowledge interacts with recall at multiple levels of abstraction. The nature of the combination of different sources of prior knowledge might depend on familiarity, where specific level prior knowledge is engaged for recall of familiar events, but prior knowledge at a higher level of abstraction is engaged for recall of unfamiliar events.
Using prior knowledge at multiple levels of abstraction is an efficient strategy that allows generalization over experiences and correction of noisy memories \cite{hemmer2009bayesian,hemmer2009integrating}.
 A similar type of knowledge transfer has long been argued to be essential in robot learning (e.g., \cite{thrun1995lifelong}).  Lifelong learning for intelligent agents, and the parallel mechanism in humans, is a significant open-issue and the subject on on-going research \cite{de2021continual,van2020brain,kudithipudi2022biological,pisupati2022challenges}.

\section{Robot-Society Dyad}
\label{sec:robotsocietydyad}

Scaling robotic solutions at a societal level will fundamentally change the technology used by society including for transportation (smart cars), health care (medical robots), infrastructure (smart buildings), restaurant industry (food prep and food delivery robots), waste management (trash and recycling robots), and warehouse logistics (sorting, picking and packing robots). 
The paradigm of human feedback affecting robot behavior scales to societal participants  affecting robot behavior. Feedback takes on a more formal structure with open issues such as: who can engage with and affect the robot systems? (i.e.\ what is their organizational role?) and when are they empowered to do so? (i.e.\ what timeline is appropriate given societal goals and constraints)  The framework of stakeholder participation includes both real-time synchronous human-in-the-loop training where robots and humans learn from and adapt to one another, as well as asynchronous engagement, where social, economic and political forces can more easily affect robot design. Asynchronous societal engagement includes discussions with stakeholders, survey research, ethnographic observations, ethical evaluations, market uptake of innovations, development of guidelines/policies by organizations and institutions, legal proceedings, the development of governmental regulations, incentives, and information \cite{andrews2006professional, scott2014institutions}. A key conjecture is that all robot systems will have unintended consequences that may only be fully discovered and quantified after initial deployment. It is critical to build with the expectation of iteration and  mechanisms must exist to  collect feedback, address issues, and re-deploy.

\subsection{Societal Theory of Mind}

The concept of Theory of Mind (ToM) where robots and/or humans predict the intent of collaborators and other entities, can be extended to a societal Theory of Interaction with Social Systems (TISS) to enable robotic systems to anticipate the needs and intentions of human groups, organizations, and institutions.
TISS models are different from traditional or machine ToM models because the fundamentals of human-robot interaction change with scale. Robot systems interacting with a community have different properties and necessitate different evaluation than the individual human-robot framework. For example, robot systems impact the culture, economy and infrastructure of a society. Predicting the mental state of humans in TISS encompasses the prediction of human groups, crowds, and assemblies. Similarly robots may act in groups or swarms and agents working with such groups need cues to predict their intent to facilitate appropriate interaction.  Computational methods for robot vision, language and action provide can provide such cues by sensing people over space and time drawing conclusions on the needs of groups. To date, both human and machine ToM models are open research issues and the problem of TISS has received sparse attention.   

%\paragraph{Structure vs. Agency}

When scaling up from interactions between individual robots and humans to those at the societal level, it becomes important to acknowledge that social, economic, and political systems exert forces on individuals and vice versa. There is a classic debate in the social sciences about whether structure constrains the agency of individuals, or whether agency determines structure.  
Contrast “class shapes everything about our lives” (Karl Marx), “we are constrained by social norms and values” (Emile Durkheim), and “bureaucracies can confine us” (Max Weber); with “individual behavior as a shaping force” (Norbert Elias), society as “the free-playing, interacting interdependence of individuals” (Georg Simmel), and “individual adaptation, goal attainment, integration and latency form the basic characteristics of social action” (Talcott Parsons). 
The message for socially cognizant roboticists is that social structures are both persistent and malleable. In the short run, social practices offer resistance to change, and humans will enforce contextualized behavioral norms on robots and one another (e.g. sidewalk delivery robots should stay navigate in a manner similar to that of humans \cite{Dalton_Montello_2019}). 
%Thus sidewalk delivery robots should stay to the right and maintain a respectful distance from other sidewalk users, whereas on roadways, autonomous vehicles should follow the specific rules of the road that apply in that jurisdiction (which includes staying to the left in the UK and Japan). 
% removed the rest to edit for size 
%These rules range from informal norms to formal, long-lasting policies or regulations. In the longer run, social groups and polities sometimes adopt new norms and policies, as when the United States switched in 1926 from a Common Law principle that “all persons have an equal right in the highway” to a (model) Uniform Vehicle Code that prioritizes motor vehicles and limits pedestrian rights (Davis 1963), or when Sweden switched from driving on the left to the right in 1967 (Raphael 1967). 

\subsection{Robots and the Built Environment}

Performance improvements in robot technology  will lead likely lead to  ubiquitous  devices  in our streets and homes. 
As industry an agencies seek to deploy robots for a myriad of tasks including food delivery, landscsaping, surveillance, advertisement and entertainment, the question will soon arise:  What will our future world look like? How many robots are too many?  We can look to the past and see the transition from nature to the built environment and learn mistakes and successes. In particular, though the field of neuroaesthetics \cite{coburn2022architectural,coburn2017buildings,chatterjee2021neuroaesthetics,karakas2020exploring,chatterjee2014neuroaesthetics} primarily focuses on architecture, it is highly relevant to the proliferation of robotics in society. This field examines the intersection of neuroscience and architectural design so that by understanding human response to the built environment (e.g.\ stress vs.\ well-being), architects can account for this response in their design. The relevance of neuroaesthetics to socially cognizant robotics is clear; the appearance of robotic systems can foster or inhibit human and societal trust of robotics. 
%The goal of introducing technology without maligning the visual habitat of humans is critically important. 
The near-term future may bring a proliferation of robots, drones and related infrastructure such as cell towers, charging stations, and storage. 
Headless dog-robots, snake robots, spider robots and other embodiments may cause stress in humans, much like a heavily industrialized urban environments do when compared to nature \cite{bratman2012impacts,hartig2014nature}. Intuitively, the effect may be amplified as the number of robots increases. 
%The importance of urban aesthetics in robot technology has  received only limited attention \cite{mladenovic2019emerging} for the specific case of self-driving cars.
The relevance of neuroaesthetics to socially cognizant robotics is clear: the appearance of robotic systems can foster or inhibit human and societal trust of robotics.

\subsection{Unintended Consequences and Policy Development}
Innovations sometimes bring unintended adverse consequences for members of society, such as birth defects in children whose mothers used the sedative Thalidomide, formation of a hole in the stratospheric ozone layer from fugitive emissions of chlorofluorocarbon refrigerants, and political violence encouraged by Facebook’s news feed algorithm. Markets can avoid obvious adverse consequences for buyers because innovations often get adopted slowly, over time, on a voluntary basis, in a bottom-up manner that allows quick feedback and decentralized responses to bad news. Political systems often react more slowly, relying on tort liability to redress harm to individuals and the (often lengthy) development of public policies to remedy wider societal harms. 

Unintended consequences have been the object of much social inquiry, for example  identification of key sources \cite{merton1936unanticipated} (ignorance, error, willful ignorance, paradoxical values, and self-defeating predictions);  
relational thinking about what is known and unknown to us and others \cite{luft1955johari}; 
and  assertations that both outcomes and likelihoods may be problematic \cite{stirling2010keep}, thus contrasting risk, uncertainty, ambiguity, and ignorance. Of particular importance in robotics are the unanticipated consequences of deployment at scale, where consequences may be economic (displaced workers), social (lonely seniors), or political (projection of military power).

It would be useful to anticipate unintended consequences of widespread deployment of robotics technologies beforehand instead of waiting for the harms to become evident. Technology assessment efforts attempt to do this by looking forward into an uncertain future using a variety of tools. These include reasoning from historical analogies, modeling socio-economic systems in order to project future impacts, fostering discussion and debate about the wisdom of deploying the innovation, reflecting on professional practices, learning from small experiments, and others \cite{Andrews2021,pringle2016unintended}. Public policies can encourage either risk-taking to spur innovation or precaution to reduce potential harms. The first may yield unintended adverse consequences and the second may stifle innovation.  Advocating {\it humility in design} can encourage the expectation of unintended consequences and the planning of design iterations.

%\subsection{Policy Development}
Public policymakers strive to encourage innovation while avoiding harm by iteratively acting and learning, a process known as “muddling through” or mutual incremental adjustment \cite{lindblom1959}. It stands in contrast to a planning or optimization approach in which decision makers assume that they already have all of the needed knowledge to make correct decisions. The incrementalist, iterative approach works best when lots of small experiments take place and there is a systematic effort to learn from the experiments. 
There is a strong movement for policy and regulation in AI (especially in generative AI), there are well known problems and cases of unregulated AI leading to unexpected consequences. There has been recent attention to regulating generative AI \cite{aiwhitehouse23,wischmeyer2020regulating,hine2023blueprint},
but less attention to robotics, although robotics can be much more directly dangerous. Self-driving cars are regulated \cite{iihs23}, but these regulations are for a narrow application area.  
%Note that Microsoft and Google have already used ChatGPT for robotics
% (https://www.microsoft.com/en-us/research/group/autonomous-systems-group-robotics/articles/chatgpt-for-robotics/)
Future regulation and policies for robotics can extend  to multiple application contexts, global collaborations \cite{salzmancollaborative}, and can
draw from core human-centric objectives as we have done for formulating best practices for socially cognizant robot design. 

%As a counterpart to Explainable AI, Explainable Robotics can be an integral part of Socially Cognizant Design in order to explain robot function to the human user. In addition, an explainable robotics framework provides a common space for robot and human to communicate; this may include an abstracted map of the environment, semantic labels of nearby objects, a sequence of subtasks for the robot action, or other interpretable and actionable representation. Deep learning based robot training can suffer from a black-box characteristic making affecting the training difficult in end-to-end model weight estimation. Explainable robotics can assist in a socially cognizant training both for the roboticist and the end-user. 

\section{Socially Cognizant Robot Design: Best Practices}
\label{sec:SCdesign}
\begin{figure}
    \centering
    \includegraphics[width=6.4in]{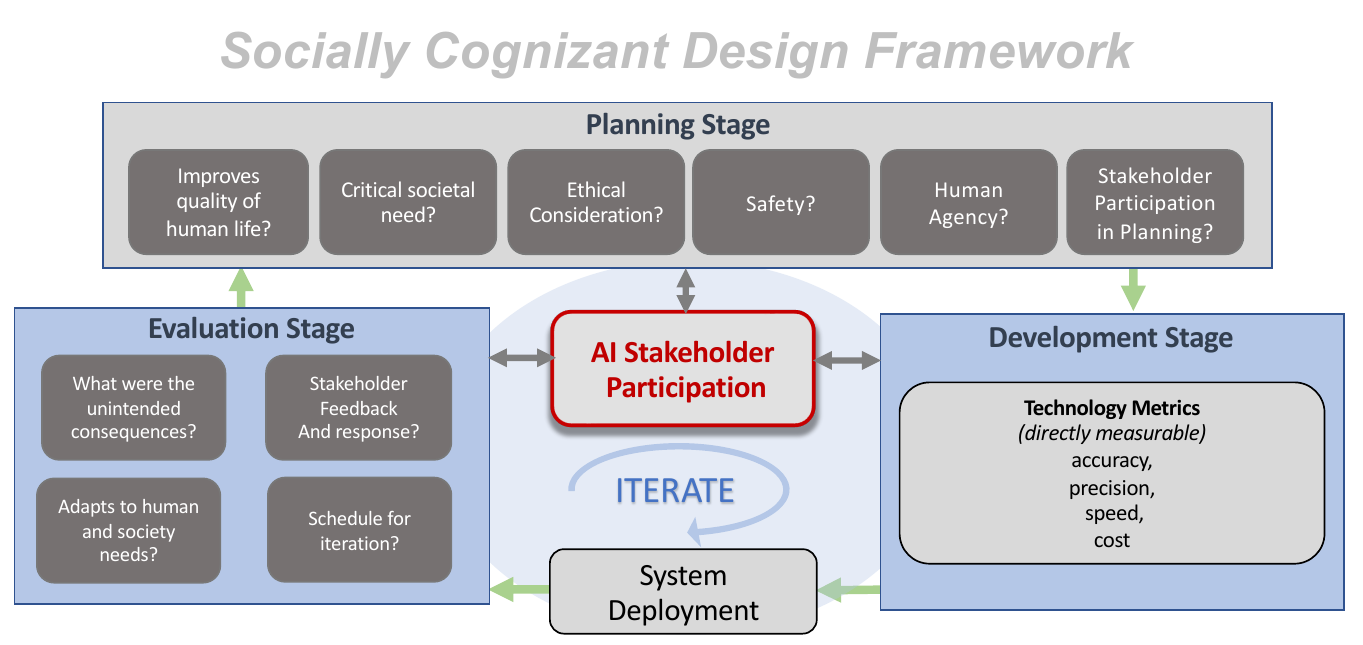}
    \caption{Standard technical design optimizes measurable metrics such as accuracy, precision and speed. Socially cognizant design is a more complex process because humanity-based metrics are not easily quantifiable but are critically important. We propose a framework that embeds meaningful best practices in the form of addressing 10 questions as described in Section~\ref{sec:SCdesign}. A key element is {\it AI stakeholder participation} both individual and societal for affecting and tuning  robot behavior. Additionally, humility in design drives the need for iterations and evaluation of each iteration. }
    \label{fig:SCdesign}
\end{figure}

Standard technical design optimizes measurable metrics such as accuracy, precision and speed. Socially cognizant design is a more complex process because humanity-based metrics are not easily quantifiable but are critically important. Guided by four core human-centric objectives described in Section~\ref{sec:intro}, we present a list of 10 questions that should be considered during a socially cognizant design. 
This proposed question set and design framework comprise best practices in socially cognizant design. As robotics advances  and the quest for socially cognizant robotics grows, we expect the enumeration of best practices to evolve further. 

\paragraph{Ten Questions to Address for Socially Cognizant Robotics Design}
\begin{enumerate}
\item Does the design improve {\bf quality of life}?
\item Does this technology address a {\bf critical societal need}? The designer should evaluate the significance given competitive technologies and consider that the technology uses limited economic and environmental resources.  
\item Have {\bf discussions} occurred with a diverse set of stakeholders prior to deployment? Consult and ask stakeholders (e.g., customers, casual users, bystanders, managing organizations, etc.) to predict the impact. 
\item What are the potential {\bf unintended consequences}? Predict unintended consequences and enumerate them before iterating over the technology. Discuss with other designers that have previous experience of robot deployment or previous users of related/competing technology. 
\item	Are there {\bf safety issues} in the application domain and have the appropriate permissions and inspections been completed? Are existing requirements sufficient to ensure human safety?
\item What is the {\bf schedule for iteration}? There is tension between speed of development and the goal of socially cognizant design. Managing this tension is related to the need for iteration.
%Planning for iteration (note: goes along with asynchronous) (note: humility in design) %Acknowledge the tension between speed of development and goal of socially cognizant design. Managing this tension is related to the need for iteration. This goes beyond the beta version
\item How will the technology {\bf adapt to human desires} and needs of people once deployed?
\item Is there a mechanism for users to provide {\bf feedback} and is there a plan to act on that feedback (asynchronous engagement)?
\item	What are the {\bf ethical considerations}?  Discuss with ethicists and groups that may be impacted by the technology.
%Enumerate ethical questions.
\item	Does the design {\bf respect human agency} and autonomy? Complete autonomy may be desirable in some setups but may also be less adaptive to human needs in others.  What are the specific mechanisms for synchronous engagement, i.e. for the human to be in the loop, to bypass automated decisions or to improve automated decisions?  Is there sufficient recognition of the limitations of automated decisions? 
\end{enumerate} 

In real-world design, while a socially cognizant solution may be desirable, budget and time constraints lead roboticists to first develop a  technology-only solution.  A meaningful step toward socially cognizant robotics is to have a deliberate design step of {\it AI stakeholder participation} to affect AI-driven robot behavior. AI stakeholders in this context are the individuals, groups,  businesses, and communities that are involved with and impacted by robotics. This participation may be asynchronous (e.g., stakeholder meetings  to understand the value of a design, user surveys to assess impact) or synchronous (e.g., robot training affected by individual or societal interaction).
Iterations of building/deploying/evaluating after multiple rounds of stakeholder assessment is crucial for meeting human-centric objectives. 
% As part of enabling people to meaningfully affect the development of robotic technology,  explainable tools for robots is a natural step.

\section{Conclusion}

Traditionally, robotics (e.g. in the context of manufacturing) has aimed primarily for speed, accuracy, and efficiency.  As robots are deployed in a wider variety of domains, it becomes important to consider human-centered objectives such as safety, adaptability, privacy, bias, and ethical considerations. At the same time, traditional social sciences have usually studied the effects of technology on society only after deployment. Given the potential impact of robotics, we cannot afford to evaluate the societal and human impact of this technology a posteriori. 

The limited exposure of robotics professionals to behavioral and social sciences makes them less than fully prepared to predict the impact of the technology they develop on society and public discourse.  Roboticists should be empowered to identify critical societal needs that robotics technology can realistically address in order to catalyze and guide meaningful convergence research. Societal needs include providing inexpensive and effective public services, efficient transportation, and environmental protection. Meeting these needs with robotic systems can transform our society by augmenting human abilities. 

Socially cognizant robotics differs from human-centered robotics by going beyond human-robot interaction to consider robot-society interactions. Furthermore, socially cognizant robotics emphasizes the importance of empowering humans to affect robot behavior both synchronously and asynchronously, recognizing the need for human agency.    It includes the objective of ethical development of robotics with the goal of enhancing life for both the individual and society. As technology increasingly infiltrates daily life, humans have become less trustful of the pervasive applications, cameras, and automation. While convenience and ease of communications has been extolled, critical failures are arising, such as biased algorithms, fatal accidents with automated vehicles, and ambiguous accountability. Socially cognizant roboticists should adhere to and contribute to the evolving ethics guidelines of automation and train scientists who consider this component at the earliest stage of robot design as well as after robot deployment.

\section{Acknowledgements}
This work has been supported by the National Science Foundation (NSF) National Research Traineeship (NRT) entitled "NRT-FW-HTF: Socially Cognizant Robotics for a Technology Enhanced Society (SOCRATES)"
Grant No. 2021628.

\bibliography{scibib}

\bibliographystyle{plain}

\section*{Acknowledgments}
This work has been supported by the National Science Foundation (NSF) National Research Traineeship (NRT) entitled "NRT-FW-HTF: Socially Cognizant Robotics for a Technology Enhanced Society (SOCRATES)"
Grant No. 2021628.
\end{document}